\documentclass[pdftex,twocolumn,epjc3]{svjour3}          

\RequirePackage[T1]{fontenc}
\let\llncssubparagraph\subparagraph
\let\subparagraph\paragraph
\usepackage[compact]{titlesec}
\let\subparagraph\llncssubparagraph
\usepackage{titlesec}
\titlespacing*{\section}{0pt}{2ex}{2ex}
\smartqed  

\RequirePackage{graphicx}
\RequirePackage{mathptmx}      
\RequirePackage{flushend}
\usepackage{graphicx}
\usepackage[colorlinks=true,linkcolor=blue,citecolor=blue]{hyperref}

\usepackage{color}

\usepackage{cite}
\usepackage{float}
\usepackage{lipsum}
\usepackage{xcolor}

\usepackage{url}
\usepackage[section]{placeins}
\usepackage{amsmath,amssymb,amsfonts}
\usepackage{algorithmic}
\usepackage{multirow}
\usepackage{multicol}
\usepackage{balance}
\usepackage{subcaption}
\usepackage{textcomp}
\usepackage{xcolor}
\usepackage{graphicx}
\usepackage{tikz}
\usepackage{pifont} 
\usepackage{bbding}

\journalname{Journal Information}

\begin{document}

\title{Deep Learning-Driven Malware Classification with API Call Sequence Analysis and Concept Drift Handling

}


\author{Bishwajit Prasad Gond\thanksref{e1,addr1}
        \and
        Durga Prasad Mohapatra\thanksref{e2,addr1} 
}

\thankstext{e1}{ORCID: \href{https://orcid.org/0000-0003-3640-0463}{0000-0003-3640-0463}}
\thankstext{e1}{e-mail: bishwajitprasadgond@gmail.com}
\thankstext{e2}{e-mail: durga@nitrkl.ac.in}

\institute{National Institute of Technology, Rourkela, Odisha, India\label{addr1}
}

\date{Received: 28$^{th}$ Feb, 2025 / Accepted: date}

\maketitle

\begin{abstract}
Malware classification in dynamic environments presents a significant challenge due to concept drift, where the statistical properties of malware data evolve over time, complicating the detection effort. To address this issue, we propose a deep learning framework enhanced with a genetic algorithm to improve malware classification accuracy and adaptability with concept drift handeling. Our approach incorporates mutation operations and fitness score evaluations within genetic algorithms to continuously refine the deep learning model, ensuring robustness against evolving malware threats. Experimental results demonstrate that this hybrid method significantly enhances classification performance and adaptability, outperforming traditional static models. Our proposed approach offers a promising solution for real-time malware classification in ever-changing cybersecurity landscapes.
\end{abstract}


\section{Introduction}\label{intro}

In the ever-evolving landscape of cybersecurity, the analysis of Portable Executable (PE) files---a prevalent format for executable programs in Windows operating systems---stands as a cornerstone for detecting and mitigating malware threats. PE files encapsulate critical structural and behavioral information, making them a prime target for malicious exploitation by cybercriminals. As malware authors continuously refine their techniques to obfuscate code, bypass signature-based detection, and exploit system vulnerabilities, traditional static analysis methods falter, unable to keep pace with the sophistication and dynamism of these threats. This necessitates a shift toward dynamic analysis approaches that can adapt to the mutable nature of malware, a challenge compounded by the phenomenon of \textit{concept drift} \cite{tsymbal2004problem}.

Concept drift, defined as the temporal alteration in the statistical properties of a target variable, poses a significant hurdle in maintaining the efficacy of machine learning models for malware detection. In the context of cybersecurity, this drift manifests as evolving malware behaviors---such as changes in API call patterns, payload delivery mechanisms, or encryption strategies---driven by adversaries’ relentless innovation. Without adaptive mechanisms, pre-trained models quickly degrade, failing to recognize new variants or emerging families of malicious software \cite{lu2018learning}. Motivated by this pressing need for adaptability, our study introduces a novel framework that synergistically combines advanced deep learning (DL) techniques with genetic algorithms (GAs) to enhance the accuracy, efficiency, and resilience of dynamic PE malware analysis.

Our proposed model leverages the power of neural networks, specifically Artificial Neural Networks (ANN), Convolutional Neural Networks (CNN) and Recurrent Neural Networks (RNNs) to classify malware samples into distinct families based on their behavioral signatures, extracted primarily through $n$ gram analysis of API call sequences. These sequences, derived from dynamic execution in sandbox environments, encapsulate the runtime interactions of PE files with the operating system, offering a rich feature set for discerning malicious intent. However, the true innovation of our approach lies in its integration of GAs to address concept drift. Inspired by natural selection, GAs enable continuous model evolution by applying mutation operations and fitness-based selection to adapt feature sets and optimize classification performance in real time \cite{vivekanandan2011mining}. This dual strategy ensures that our system not only excels in categorizing known malware but also dynamically adjusts to emerging threats, a capability critical for safeguarding systems in today’s volatile cyber landscape.

The motivation behind this research stems from the escalating arms race between malware developers and cybersecurity defenders. Static models, while computationally efficient, lack the flexibility to counter the rapid evolution of malware, as evidenced by studies highlighting performance degradation over time \cite{chen2023overkill}. Conversely, purely DL-based approaches, though powerful in pattern recognition, often struggle with generalization across drifting data distributions without retraining \cite{jameel2020critical}. By marrying DL’s robust feature extraction with GAs’ evolutionary adaptability, we aim at bridging this gap, offering a proactive, scalable solution that anticipates and responds to shifts in malware characteristics. This work builds on prior efforts to handle concept drift \cite{garcia2023effectiveness, fernando2024fesad} while pushing the boundaries of malware classification through a focus on API call sequences---a dynamic, behavior-driven indicator of malicious activity.

Our objectives are twofold: first, to harness deep learning techniques to uncover intricate patterns and behaviors within malware samples, thereby improving classification precision; and second, to enhance adaptability to concept drift, ensuring long-term effectiveness against sophisticated cyber threats. To this end, the paper explores the application of DL and GA methodologies for dynamic malware analysis, with a particular emphasis on $n$-gram API call analysis as a lens into malware behavior.

The remainder of this paper is structured as follows: Section \ref{sec:basicc} elucidates the foundational concepts of concept drift, its variants, and the pivotal role of GAs and sandbox environments in addressing it. Section \ref{sec:related} reviews the related work, situating our contribution within the broader research landscape. Section \ref{sec:datacollection} delineates the data collection and preprocessing pipeline critical to our experiments. Section \ref{sec:cdarch} presents our integrated DL and GA-based framework, detailing its phases from preprocessing to concept drift handling. Section \ref{sec:cdimple} describes the experimental setup---including API key acquisition, data extraction, and model training---and provides an in-depth analysis of results across diverse DL algorithms paired with GAs. Section \ref{sec:cdcom} benchmarks our approach against state-of-the-art techniques, highlighting its strengths and novelty. Section \ref{sec:threat} discusses threats to validity, ensuring a rigorous evaluation of our findings. Finally, Section \ref{sec:conclusion} summarizes our contributions and outlines avenues for future research, reinforcing the transformative potential of our methodology in fortifying cybersecurity defenses.

\section{Foundational Concepts for Malware Analysis}\label{sec:basicc}

This section lays the groundwork for understanding the core methodologies employed in our study by elucidating the essential concepts and terminologies central to dynamic malware analysis. We focus on three intertwined pillars: \textit{concept drift}, which captures the evolving nature of malware data; \textit{techniques to handle concept drift}, which ensure model robustness; and \textit{genetic algorithms}, which provide an adaptive framework for addressing these shifts. These concepts are critical to our proposed deep learning (DL) and genetic algorithm (GA)-based approach for classifying Portable Executable (PE) malware, particularly in the context of API call sequence analysis.

\subsection{\textbf{Concept Drift: Definition and Taxonomy}}
Concept drift refers to the phenomenon wherein the statistical properties of a target variable---the entity a predictive model aims to classify or forecast---undergo unforeseen changes over time \cite{tsymbal2004problem, lu2018learning}. In machine learning, this drift manifests as a discrepancy between the training data distribution \( P(X, y) \) and the test-time distribution \( P'(X, y) \), where \( X \) represents input features (e.g., API call sequences) and \( y \) denotes the target labels (e.g., malware families). Formally, concept drift occurs when:
\[
P_t(X, y) \neq P_{t+1}(X, y),
\]
where \( t \) and \( t+1 \) denote successive time steps. In the cybersecurity domain, such shifts are driven by factors like adversarial innovation (e.g., new obfuscation techniques), environmental changes (e.g., operating system updates), or data source variations (e.g., emerging malware strains). For malware detection, concept drift complicates the maintenance of model accuracy, as static models trained on historical data fail to generalize to novel threats.

To characterize this phenomenon, concept drift is categorized into distinct types based on the nature and tempo of the change:
\begin{enumerate}
    \item \textbf{Sudden Concept Drift}: This occurs when the target distribution shifts abruptly, often triggered by a discrete event. For instance, the release of a new malware variant (e.g., WannaCry ransomware in 2017) could abruptly alter the distribution of malicious behaviors, rendering existing models obsolete overnight. Mathematically, this is a step change: \( P_t(y|X) \to P_{t+1}(y|X) \) where \( P_{t+1} \) differs significantly within a short timeframe.

    \item \textbf{Incremental Concept Drift}: This type involves a gradual, continuous evolution of the target distribution. An example is the slow adaptation of trojan malware to employ more sophisticated encryption over months, making detection incrementally harder. This can be modeled as a smooth transition: \( P_t(y|X) \to P_{t+1}(y|X) \) with small, cumulative changes.

    \item \textbf{Recurring Concept Drift}: Here, the distribution exhibits cyclic or periodic patterns. Seasonal ransomware campaigns, such as those peaking during tax seasons, exemplify this, where \( P_t(y|X) \) reverts to a prior state after a cycle (e.g., \( P_{t+n}(y|X) \approx P_t(y|X) \)) \cite{suarez2023survey}. This requires models to recognize and adapt to temporal recurrence.

    \item \textbf{Concept Drift by Context}: This arises when the relationship between inputs \( X \) and outputs \( y \) varies with contextual factors. For malware, context might include the target operating system version; a backdoor’s behavior may differ between Windows 7 and Windows 10, necessitating context-aware modeling.

    \item \textbf{Covariate Shift}: Unlike other types, covariate shift involves a change in the input distribution \( P(X) \) while the conditional distribution \( P(y|X) \) remains stable \cite{nair2019covariate}. In malware analysis, this might occur when API call frequencies shift due to software updates, yet the underlying malicious intent remains consistent. Mitigation often involves reweighting samples via importance sampling: \( w(X) = \frac{P_{\text{new}}(X)}{P_{\text{old}}(X)} \).

\end{enumerate}

Detecting and adapting to concept drift is paramount are dynamic environments like malware detection, where failure to adapt leads to degraded performance (e.g., increased false negatives). The existing techniques range from passive monitoring of prediction errors to proactive model updates, as detailed in the following subsection.

\subsection{\textbf{Techniques to Handle Concept Drift}}
Maintaining model efficacy amidst concept drift requires a suite of adaptive strategies tailored to the drift’s characteristics. Below, we outline the key techniques, emphasizing their applicability to malware classification:
\begin{enumerate}
    \item \textbf{Re-training}: Periodic retraining updates the model with fresh data, resetting parameters to reflect \( P_{t+1}(X, y) \). For sudden drift (e.g., a new worm outbreak), full retraining on recent samples restores accuracy, while incremental drift benefits from partial updates. However, this approach demands computational resources and timely data availability.

    \item \textbf{Ensemble Methods}: Techniques like bagging and boosting combine multiple models (e.g., decision trees, neural networks) trained on diverse data subsets or time windows. For malware, an ensemble might include classifiers for old and new variants, weighted by recency or performance, enhancing robustness to recurring drift \cite{kuncheva2004classifier}.

    \item \textbf{Online Learning}: This enables continuous parameter updates as new data arrives, ideal for incremental drift. Algorithms like stochastic gradient descent (SGD) adjust weights incrementally: \( w_{t+1} = w_t - \eta \nabla L(x_t, y_t) \), where \( \eta \) is the learning rate and \( L \) is the loss. In cybersecurity, this suits streaming malware data from live systems.

    \item \textbf{Change Detection Algorithms}: Methods such as Cumulative Sum (CUSUM) or Page-Hinkley test monitor performance metrics (e.g., error rate) for drift signals. CUSUM tracks cumulative deviations: \( S_t = \max(0, S_{t-1} + (x_t - \mu)) \), triggering retraining when \( S_t \) exceeds a threshold. This is vital for sudden drift detection in malware campaigns.

    \item \textbf{Instance Weighting}: Techniques like Importance Weighted Cross-Validation (IWCV) assign higher weights to recent instances, adapting to incremental drift. For a sample \( (x_i, y_i) \), the weight \( w_i = e^{-\lambda (t - t_i)} \) prioritizes newer data, where \( \lambda \) controls decay rate.

    \item \textbf{Feature Selection and Extraction}: Dynamically updating features (e.g., selecting new API calls) aligns the model with shifting distributions, critical for covariate shift in malware datasets.

    \item \textbf{Windowing}: A sliding window trains on the most recent \( n \) samples, discarding older data. For gradual drift, a fixed-size window (e.g., 1000 samples) ensures relevance, though it risks losing long-term patterns.

    \item \textbf{Dynamic Feature Adaptation}: This modifies feature representations (e.g., $n$-gram weights) based on current data, enhancing pattern capture amidst drift.

    \item \textbf{Memory-Based Methods}: Approaches like k-Nearest Neighbors (k-NN) store historical data, querying it for predictions. Adaptive k-NN adjusts the neighbor set dynamically, balancing memory and recency for recurring drift.

\end{enumerate}

Each strategy offers trade-offs: online learning excels in real-time settings but may overfit noise, while ensembles provide stability at higher computational cost. The choice depends on drift type, data volume, and application constraints, Our study uses GAs for their evolutionary adaptability.

\subsection{\textbf{Genetic Algorithms for Concept Drift}}
Genetic algorithms (GAs), inspired by Darwinian evolution, are optimization heuristics that emulate natural selection through mutation, crossover, and selection \cite{vivekanandan2011mining}. In machine learning, GAs address concept drift by evolving solutions to align with changing environments. Operating on a population of candidate solutions (e.g., feature sets, models), GAs iterate as follows:
\begin{enumerate}
    \item \textit{Initialization}: Generate a random population.
    \item \textit{Fitness Evaluation}: Score individuals using a fitness function (e.g., classification accuracy).
    \item \textit{Selection}: Favor high-fitness individuals (e.g., via tournament selection).
    \item \textit{Crossover}: Combine parent solutions to produce offspring.
    \item \textit{Mutation}: Introduce random changes to maintain diversity.
    \item \textit{Replacement}: Update the population with new generations.
\end{enumerate}
For handelling concept drift, GAs adapt models dynamically as follows:
\begin{enumerate}
    \item \textbf{Feature Selection}: GAs identify the relevant features (e.g., API $n$-grams) for the current drift state, evolving a chromosome of binary feature masks (1 = selected, 0 = discarded). Fitness might be the F1-score on recent data.

    \item \textbf{Model Selection}: GAs evolve model architectures (e.g., neural network layers) or types (e.g., CNN vs. RNN), optimizing for current \( P(y|X) \).

    \item \textbf{Ensemble Generation}: GAs create diverse model ensembles, each tailored to a drift phase, combining predictions via weighted voting.

    \item \textbf{Hyperparameter Optimization}: GAs tune parameters (e.g., learning rate \( \eta \), dropout rate) to maximize performance, encoding them as real-valued genes.

    \item \textbf{Instance Selection}: GAs select representative historical instances, reducing noise and aligning training data with \( P_{t+1}(X, y) \).

\end{enumerate}

In malware analysis, GAs excel by mutating API call features to mimic evolving threats (e.g., new system call patterns), ensuring robustness. For example, a fitness function \( F = \text{Accuracy} + \alpha \cdot \text{Stability} \) (where \( \alpha \) balances performance and drift resilience) guides evolution. This adaptability distinguishes GAs from static retraining, making them integral to our framework, as detailed in Section \ref{sec:cdarch}.

\subsection{\textbf{Sandboxing in Malware Analysis}}\label{sec:sandbox}

In the domain of cybersecurity and software engineering, a sandbox serves as a pivotal security mechanism designed to isolate and execute untrusted or potentially malicious programs in a controlled environment, safeguarding the underlying system from harm. Drawing an analogy from a child’s sandbox---where activities are confined to a safe, bounded space---this technique creates a virtualized or restricted execution context that prevents rogue processes from interacting directly with critical system resources \cite{vasilescu2014practical}. Sandboxes are widely employed in software development, testing, and, most notably, malware analysis, where they enable the safe detonation and observation of suspicious code. This isolation is particularly crucial for analyzing Portable Executable (PE) files, which are frequent vectors for malware due to their executable nature in Windows environments.

The operational foundation of a sandbox rests on a multi-faceted approach to resource management, monitoring, and threat containment, ensuring both security and analytical depth. First, \textbf{resource control} is enforced by imposing strict limits on the resources available to the sandboxed process. These restrictions encompass file system access (e.g., read/write permissions confined to a virtual directory), network connectivity (e.g., blocking outbound traffic or rerouting it to a honeypot), memory allocation (e.g., capping usage to prevent buffer overflows), and CPU utilization (e.g., throttling to mitigate denial-of-service risks). By constraining these resources, the sandbox minimizes the potential damage that malicious code could inflict, even if it attempts to exploit vulnerabilities \cite{afianian2019malware}.

Second, \textbf{monitoring and analysis} form the backbone of a sandbox’s diagnostic capabilities. Advanced sandboxes are equipped with instrumentation to log and scrutinize a wide array of system interactions, including system calls (e.g., \texttt{NtCreateFile}, \texttt{NtWriteFile}), network activity (e.g., DNS queries, TCP/UDP packet flows), file operations (e.g., creation, modification, deletion), and registry manipulations (e.g., key creation or value changes). These logs are analyzed in real time or post-execution to identify behavioral signatures indicative of malice, such as attempts to escalate privileges, inject code into other processes, or establish command-and-control (C2) communications. This dynamic analysis contrasts with static analysis by capturing runtime behavior, offering a more comprehensive view of a malware sample’s intent and capabilities \cite{jamalpur2018dynamic}.

Third, the sandbox employs \textbf{dynamic analysis} to detect and respond to threats as they unfold. Unlike static methods that disassemble code without execution, dynamic analysis executes the sample within the sandbox, observing its interactions with the emulated system. Techniques such as API hooking, kernel-level monitoring, and behavioral heuristics enable the identification of exploits (e.g., zero-day vulnerabilities) or malicious actions (e.g., ransomware encryption routines). This real-time scrutiny is computationally intensive but invaluable for uncovering obfuscated or polymorphic malware that evades static detection \cite{afianian2019malware}.

Finally, \textbf{containment} mechanisms ensure that detected threats are neutralized effectively. Upon identifying suspicious behavior---such as a process attempting to overwrite the Master Boot Record (MBR) or connect to a known malicious IP---the sandbox can terminate the process, roll back system changes using snapshot restoration (common in virtualized sandboxes), or quarantine the sample for further study. This containment not only prevents harm but also preserves the integrity of the analysis environment, allowing repeated executions under varied conditions.

A prominent implementation of these principles is \textbf{Cuckoo Sandbox}, an open-source, automated malware analysis system accessible at \url{https://github.com/cuckoosandbox/} \cite{cuckoo_repo}. Cuckoo Sandbox exemplifies the sandbox paradigm by providing a robust framework for analyzing suspicious files and URLs within a virtualized setting. It leverages virtualization technologies (e.g., VirtualBox, VMware, or KVM) to instantiate isolated guest environments---typically Windows instances---where PE files are executed safely. Cuckoo’s architecture comprises of a host system that orchestrates analysis and a guest virtual machine (VM) where the malware is detonated. During execution, it captures detailed telemetry, including API call sequences, network traffic (via packet capture), and system modifications, which are compiled into comprehensive JSON reports for subsequent investigation \cite{jamalpur2018dynamic}. These reports are instrumental in our study, as they provide the $n$-gram API call data driving our deep learning-based classification.

To deploy Cuckoo Sandbox effectively, specific system requirements must be met. The host operating system should be a Linux distribution (e.g., Ubuntu 18.04 LTS, Debian, or CentOS) due to its stability and compatibility with virtualization tools. Hardware demands include a multi-core CPU (e.g., Intel i7 or equivalent) to support concurrent VM execution, ample RAM (minimum 8 GB, recommended 16 GB or more) to accommodate multiple guests, and substantial disk space (at least 500 GB, ideally SSD-backed) to store malware samples, VM images, and analysis logs. Virtualization software such as VirtualBox, VMware Workstation, or KVM is mandatory to create the isolated execution environments, with configuration options to emulate diverse Windows versions (e.g., Windows 10, 7) matching target malware’s operational context. Additionally, Python 3.12 is required, along with dependencies like \textbf{libvirt} and \textbf{volatility}, to support Cuckoo’s scripting and memory forensics capabilities. Proper network configuration, including a virtual NIC bridged to a monitored subnet, ensures safe capture of network-based behaviors without risking external exposure.

Cuckoo Sandbox’s strengths lie in its modularity and extensibility. It supports plugins for custom analysis (e.g., memory dump parsing with Volatility), integrates with external services like VirusTotal for signature verification, and scales to handle large sample volumes via distributed setups. However, its reliance on virtualization introduces potential weaknesses, such as detection by VM-aware malware that alters behavior when sandboxed (e.g., checking for hypervisor artifacts like CPUID instructions). To mitigate this, advanced configurations employ bare-metal sandboxes or anti-evasion techniques (e.g., randomizing system artifacts), though these fall beyond our current scope.

In our research, Cuckoo Sandbox serves as the linchpin for dynamic PE malware analysis, enabling the extraction of runtime behaviors critical to our deep learning and genetic algorithm framework. By isolating and dissecting malware in this controlled environment, we obtain the raw data---API call sequences---that fuel our $n$-gram feature extraction, classification, and concept drift handling, as elaborated in subsequent sections.

\section{Related Work}\label{sec:related}

Chen et al. \cite{chen2023overkill} distinguished between feature-space drift and data-space drift in malware detectors, highlighting the predominant influence of data-space drift on model degradation over time. Their findings underscored the necessity for further exploration into the implications of feature-space updates, particularly in the context of Android malware datasets like AndroZoo and EMBER.

Jameel et al. \cite{jameel2020critical}  conducted a critical review of concept drift's adverse effects on machine learning classification models. They proposed the ACNNELM model as optimal for Big Data stream classification but noted the absence of critical parameters for advanced ML models like deep learning. The review also highlighted the lack of a matrix model to measure adaptability factors, suggesting avenues for future research in model evaluation and optimization.

Hashmani et al. \cite{hashmani2020concept} presented a systematic literature review on concept drift evolution in machine learning approaches. Their comprehensive synthesis and categorization of existing research provided valuable insights into the state of the art in handling concept drift. However, the paper itself did not contribute new methodologies, serving primarily as a reference for researchers seeking to understand current trends and challenges in the field.

Lu et al. \cite{lu2018learning} conducted an extensive review focusing on concept drift in machine learning, covering detection, understanding, and adaptation strategies across numerous studies. While offering valuable insights into the breadth of research in this area, the paper did not introduce novel methodologies, functioning primarily as a compilation and analysis of existing approaches.

Farid et al. \cite{farid2013adaptive} proposed an adaptive ensemble classifier for mining concept drifting data streams. Their methodology addressed the challenge of concept drift by leveraging an ensemble approach, demonstrating promise for real-world applications with evolving data streams. However, the effectiveness of their approach was contingent upon the selection of appropriate base classifiers, highlighting a potential area for improvement in future research.

\section{Data Collection and Preprocessing}\label{sec:datacollection}
\begin{figure}[h]%
    \centering
{{\includegraphics[angle=90,origin=c, width=6cm]{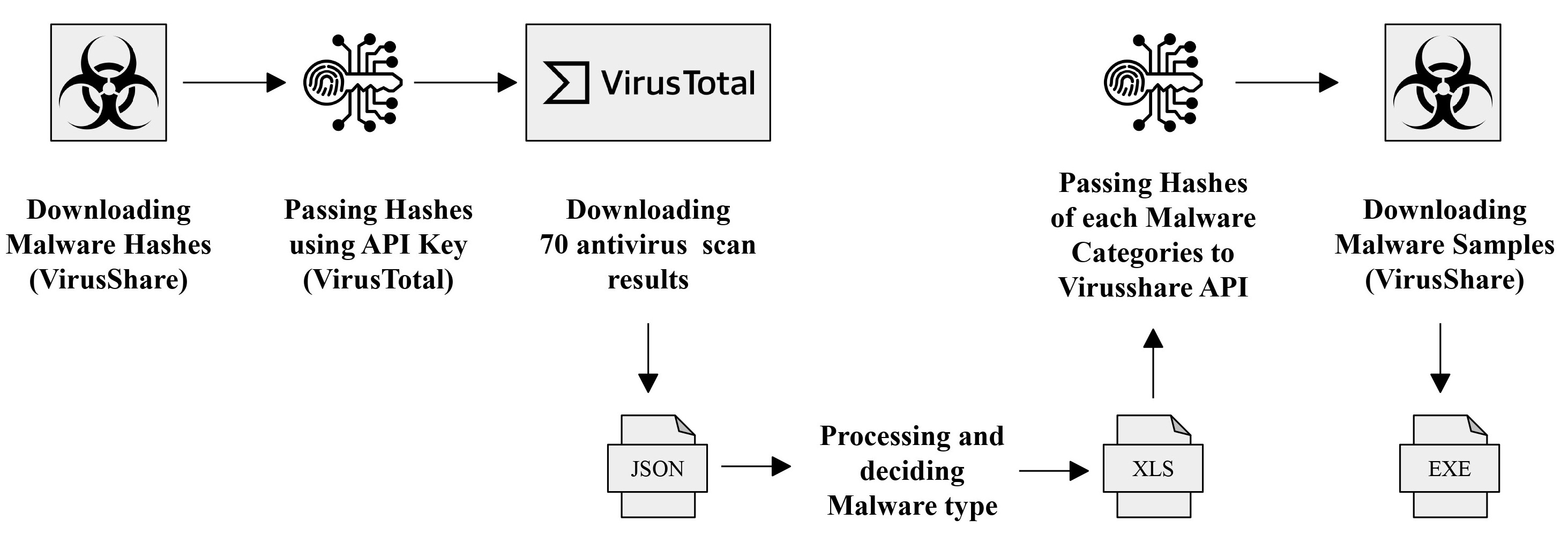} }}%
    \centering
    \caption{Data Collection Process}
    \label{fig:pearch}%
\end{figure}

In this section, we discuss the process of data collection from VirusShare \cite{virusshare} and VirusTotal \cite{virustotal}, as shown in Figure ~\ref{fig:pearch}.
\begin{enumerate}
  \item \textbf{Downloading Malware Hashes:} The process starts with obtaining malware hashes from VirusShare \cite{virusshare}, a platform known for its collection of malware samples.
  
  \item \textbf{Passing Hashes to VirusTotal:} These hashes are then sent to VirusTotal \cite{virustotal} using an API key. VirusTotal analyzes files and URLs, stores the hashes, and signatures in their repository to detect malicious content.

  \item \textbf{Downloading Antivirus Scan Results:} Subsequently, the results from 70 different antivirus scans are downloaded for each hash from VirusTotal in JSON format.

  \item \textbf{Categorizing Malware:} The JSON results are analyzed and categorized into different malware types such as adware, backdoor, trojan, spyware, virus, downloader, and worms. These categories are stored in seven different files initially. The antivirus scan results are stored in JSON format, and after processing, the malware types are documented in XLS format.

  \item \textbf{Passing Results Back to VirusShare's API:} The categorized results are then passed back to VirusShare's API.
  
  \item \textbf{Downloading Malware Samples:} Based on the categorized hashes, specific malware samples are downloaded from VirusShare.
\end{enumerate}

\section{Proposed Architecture}\label{sec:cdarch}
\begin{figure*}[h]%
\centering
{{\includegraphics[width=16cm]{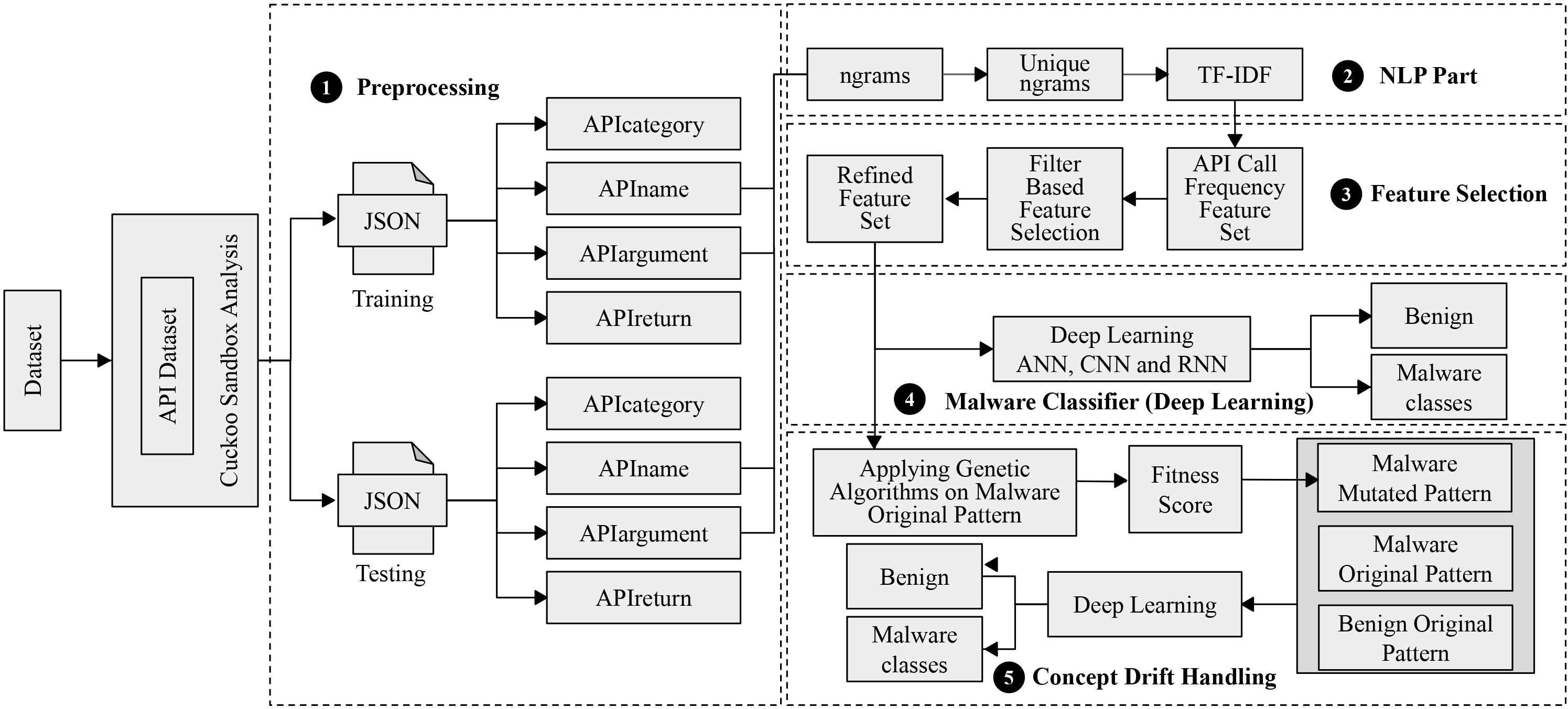} }}%
    \caption{Proposed Architecture for Malware Analysis with Concept Drift handling}
    \label{fig:cd}%
\end{figure*}

In this section, we discuss the architecture and methodology of our proposed model. The proposed model is shown in Figure \ref{fig:cd}.
The detailed design of the proposed neural network architectures is presented here. Our approach consists of five phases. Below, we explain these phases in detail.

\subsection*{\textbf{Phase 1: Preprocessing Phase}}
In the preprocessing phase, we perform the following activities.
\begin{enumerate}
    \item \textbf{API Dataset \& Cuckoo Sandbox Analysis:} The process starts with a dataset that undergoes this analysis.
    \item \textbf{Data Division:} The data is divided into training and testing sets, both processed through JSON format.
    \item \textbf{API Elements Extraction:} Various API elements like APICategory, APIName, APIArgument, and APIreturn are extracted from the API call sequence as shown in Figure \ref{fig:cuckooana}.
\end{enumerate}
\subsection*{\textbf{Phase 2: NLP Phase}}
In NLP phase, we perform the following activities.
\begin{enumerate}
    \item \textbf{Creating $n$-grams and Unique $n$-grams:} We extracted $n$-grams from the Cuckoo report (JSON). From these $n$-grams, we created unique unigram, bigram, and trigram corpora for each class of malware and benign samples.
      \item \textbf{$n$-grams after Processing:} The following are the examples of unigrams, bigrams, and trigrams that we have used in this paper.
\begin{itemize}
    \item \textbf{Unigram: }\textit{LdrLoadDll\_urlmon\_urlmon.dll}
    \item \textbf{Bigram: }\textit{NtAllocateVirtualMemory\_na,\\ LdrLoadDll\_ole32\_ole32.dll}
    \item \textbf{Trigram: }\textit{LdrUnloadDll\_SHELL32,\\ LdrLoadDll\_SETUPAPI\_SETUPAPI.dll,\\ LdrGetProcedureAddress\_ole32\_OleUninitialize}
\end{itemize}

    \item \textbf{Calculating TF:} Term Frequency is applied to transform the text data. It tokenizes text, counts the occurrences of each token, and computes TF weights. These weights reflect the importance of each token in a document relative to the entire corpus.
    \item \textbf{Refining Feature Set:} A refined feature set is obtained after filtering based on frequency.
\end{enumerate}
\subsection*{\textbf{Phase 3: Feature Selection Phase}}
In the feature selection phase, we perform the following activities.
    
\begin{figure}[htbp]%
    \centering
{{\includegraphics[angle=270,origin=c, width=6cm]{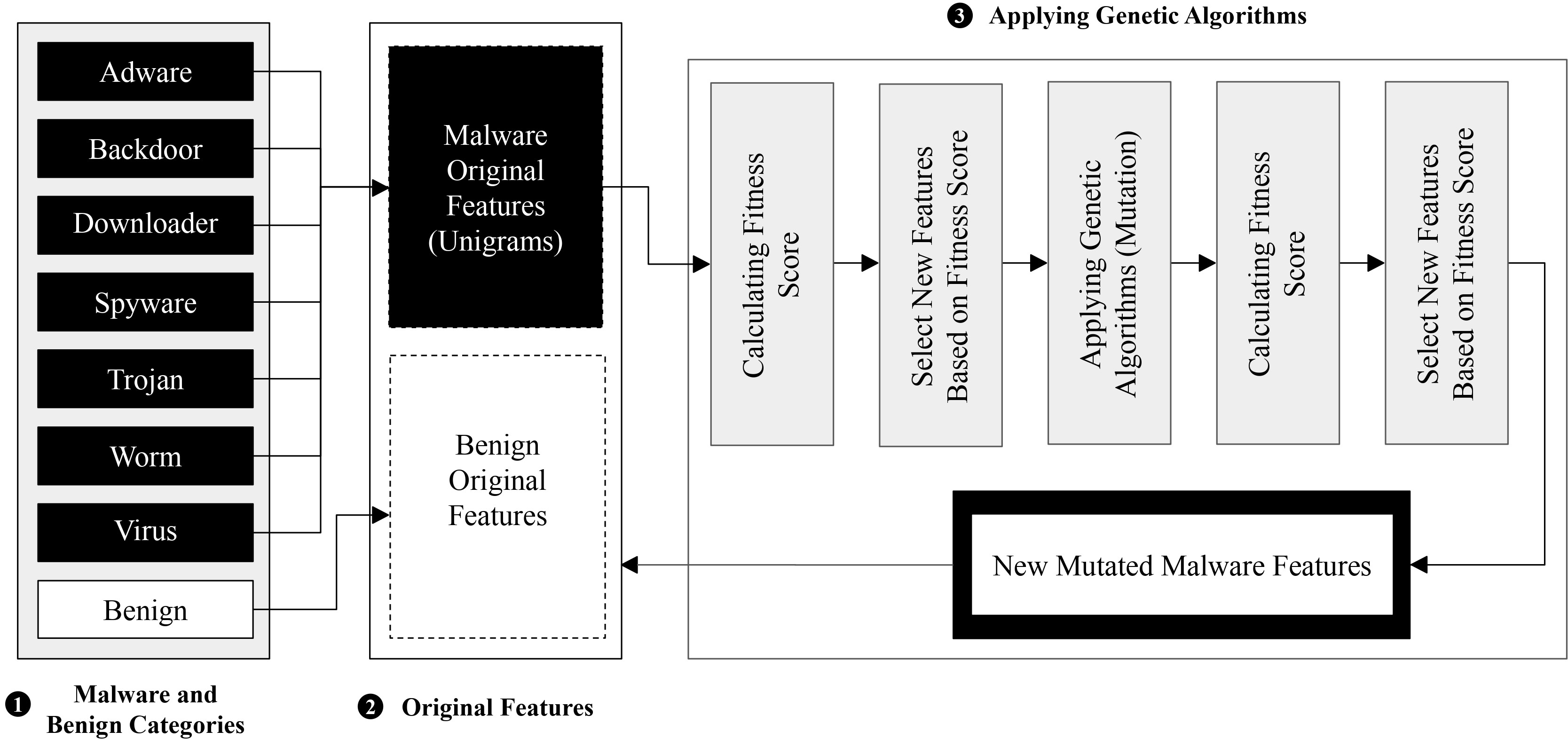} }}%
    \centering
    \caption{Mutated Features Creation using Genetic Algorithms}
    \label{fig:ga}%
    \end{figure}
\begin{enumerate}
    \item \textbf{Creating API Call Frequency Feature Set:} Explore features derived from API call frequency to understand the system behaviour and usage patterns.
    \begin{figure*}[h]%
    \centering
    {
        {\includegraphics[angle=90,origin=c, width=16cm]{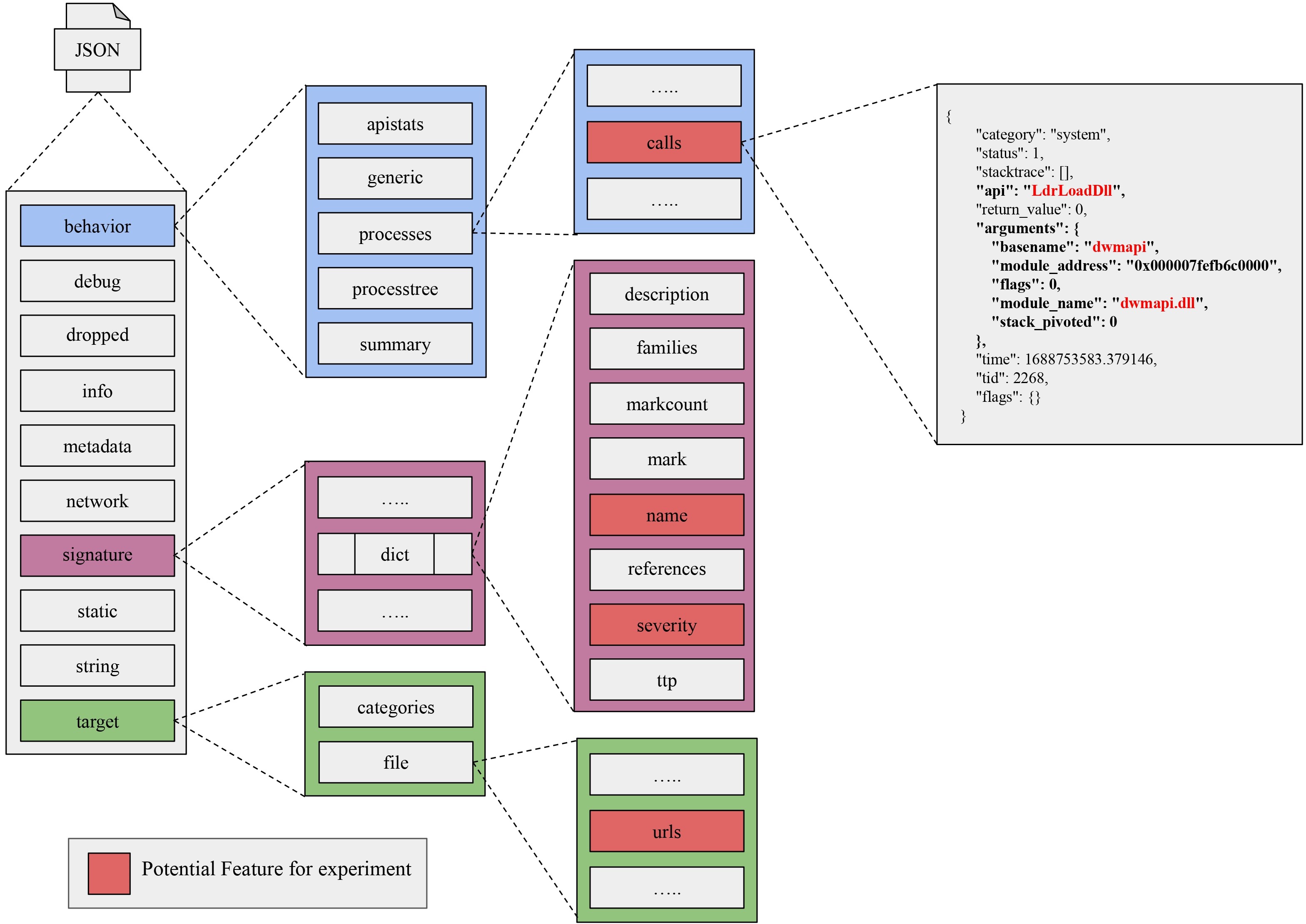} }
    }%
    \centering
    \caption{Potential Feature Selection after Cuckoo Analysis }
    \label{fig:cuckooana}%
\end{figure*}

    \item \textbf{Applying Filter Based Feature Selection:} 
    Apply filter-based techniques (e.g., mutual information, correlation analysis) to select the most informative features.
    \item \textbf{Refining Feature Set:} Eliminate redundant or irrelevant features using set hybrid feature selection techniques to ensure that the final set is discriminative and predictive.
\end{enumerate}

\subsection*{\textbf{Phase 4: Malware Classification Phase}}
In the malware classification phase, we perform the following activities.
\begin{enumerate}
    
    \item \textbf{Applying Deep Learning Technique:} In this context, deep learning techniques such as ANN, RNN and CNN are used for malware classification. Deep learning techniques involve the use of neural networks with multiple layers to learn complex patterns and representations    
    from data. This approach is well-suited for tasks such as malware classification, where the data may have intricate patterns that are difficult to capture using traditional machine-learning algorithms.
\end{enumerate}

\subsection*{\textbf{Phase 5: Concept Drift Handling Phase}}
In the concept drift handling phase, we perform the following activities.

\begin{enumerate}
   \item \textbf{Applying Genetic Algorithms on Malware Original Pattern:} Genetic algorithms are used to evolve and optimise solutions to a problem, mimicking the process of natural selection. In the context of malware analysis, genetic algorithms can be applied to evolve or mutate features of malware samples to explore different characteristics and improve classification or analysis results. The mutation occurs in the sub-parts of API sequence features of secondary and tertiary, but the primary remains untouched, as shown in Figure \ref{fig:ga} and \ref{fig:mutated}. We have generated 101248 new mutants by applying crossover and mutation on the malware's original pattern (Unigrams).
\begin{figure}[H]
    \centering
    \includegraphics[width=5.3cm]{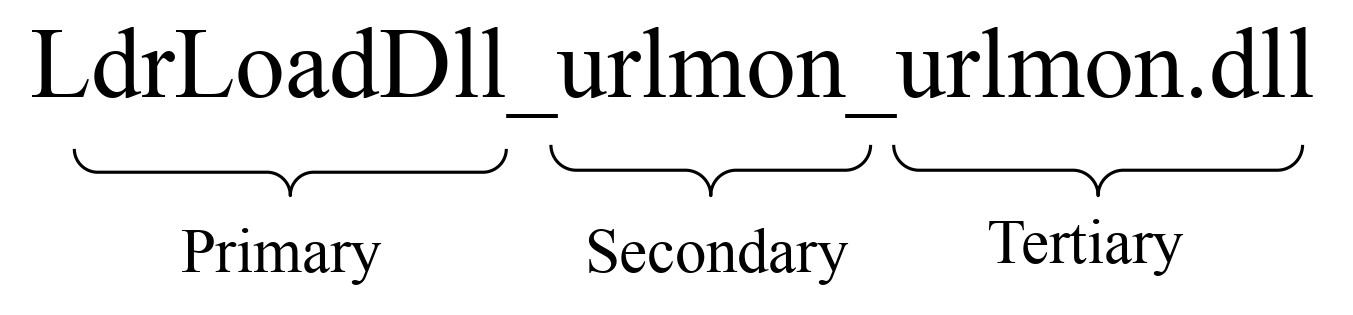}
    \caption{Mutated Features}
    \label{fig:mutated}
\end{figure}

\item \textbf{Calculation of Fitness Score:}
The fitness score \cite{kozeny2015genetic} is calculated as the edit distance from the target string. The formula for calculating the fitness score for an individual in the population is:

\begin{equation}
  \text{{Fitness}}(individual) = \sum_{i=1}^{n} 1_{a_i \neq b_i}  
\end{equation}

Where:
\begin{itemize}
    \item \( \text{{Fitness}}(individual) \) is the fitness score of the individual. 
    \item \( n \) is the length of the target string.
    \item \( a_i \) is the \( i \)-th character of the target string.
    \item \( b_i \) is the \( i \)-th character of the individual.
\end{itemize}

 This formula calculates the number of positions in the individual where the character does not match the corresponding character in the target string, summing them up to get the total edit distance. We conducted fitness score calculations on a dataset comprising 101,248 new mutants. From this pool, we selected 10,500 mutants, with the top 1500 from each malware category based on their fitness scores as shown in Table \ref{tab:fitness}. These mutants serve as features, adding approximately $\approx$ 1\% to our existing feature corpus. 
\begin{table}[h!]
\caption{Selecting new features using Fitness Score}
\begin{center}

\begin{tabular}{@{\extracolsep{\fill}}lrrrrr@{}}
\hline
\textbf{Malware} & \textbf{F$_1$} & \textbf{F$_2$}  & \textbf{...} & \textbf{F$_n$} & \textbf{Top 1500 Features} \\ \hline
\hline

Adware & 541 & 1067  & ... & 1354 & F$_1$, F$_5$, F$_9$ ... \\ \hline
Backdoor & 1265 & 1397  & ... & 1289 & F$_{85}$, F$_{1101}$, F$_{2077}$ ... \\ \hline
Downloader & 1753 & 824  & ... & 1250 & F$_{651}$, F$_{1742}$, F$_{222}$ ...\\ \hline
Spyware & 891 & 1925  & ... & 1065 & F$_{2089}$, F$_{437}$, F$_{901}$ ... \\ \hline
Trojan & 1744 & 1129  & ... & 1163 & F$_{1600}$, F$_{2200}$, F$_{882}$ ... \\ \hline
Worm & 1579 & 1367  & ... & 1625 & F$_{785}$, F$_{2023}$, F$_{1488}$ ... \\ \hline
Virus & 903 & 1064  & ... & 1293 & F$_{33}$, F$_{1999}$, F$_{777}$ ... \\ \hline
\end{tabular}
\label{tab:fitness}
\end{center}
\end{table}
\begin{table}[htbp]
\caption{CNN Model Architecture}
\begin{center}
\begin{tabular}{@{\extracolsep{\fill}}lr@{}}
\hline
\textbf{Layer} & \textbf{Details} \\
\hline
Input Shape & (88972, 1) \\
\hline
Conv. Layer 1 & Filters = 64, Kernel Size = 3, Activation = ReLU \\
\hline
MaxPooling Layer 1 & Pool Size = 2 \\
\hline
Conv. Layer 2 & Filters = 32, Kernel Size = 3, Activation = ReLU \\
\hline
MaxPooling Layer 2 & Pool Size = 2 \\
\hline
Flatten Layer & Flatten the input to a 1D array \\
\hline
Dense Layer 1 & Neurons = 128, Activation = ReLU \\
\hline
Dropout Layer 1 & Dropout Rate = 0.3 \\
\hline
Dense Layer 2 & Neurons = 64, Activation = ReLU \\
\hline
Dropout Layer 2 & Dropout Rate = 0.3 \\
\hline
Output Layer & Neurons = 8, Activation = Softmax \\
\hline
\end{tabular}
\label{tab:cnn_model_architecture}
\end{center}
\end{table}

\item \textbf{Final Feature Selection using Genetic Algorithm:} This phase involves the selection of the final feature set, which consists of the original malware set and the mutated features of malware, as well as benign features. This selection process is crucial for ensuring that the features used for analysis or classification are relevant and effective in distinguishing between malware and benign samples.

\begin{table}[htbp]
\caption{RNN Model Architecture}
\begin{center}
\begin{tabular}{@{\extracolsep{\fill}}lr@{}}
\hline
\textbf{Layer} & \textbf{Details} \\
\hline
Input Shape & (88972, 1) \\
\hline
SimpleRNN Layer & Units = 128, Activation = ReLU \\
\hline
Dropout Layer 1 & Dropout Rate = 0.5 \\
\hline
Dense Layer 1 & Neurons = 64, Activation = ReLU \\
\hline
Dropout Layer 2 & Dropout Rate = 0.5 \\
\hline
Output Layer & Neurons = 8, Activation = Softmax \\
\hline
\end{tabular}
\label{tab:rnn_model_architecture}
\end{center}
\end{table}

\begin{table}[htbp]
\caption{ANN Model Architecture}
\begin{center}
\begin{tabular}{@{\extracolsep{\fill}}lr@{}}
\hline
\textbf{Layer} & \textbf{Details} \\
\hline
Input Shape & (88972, 1) \\
\hline
Dense Layer 1 & Neurons = 512, Activation = tanh \\
\hline
Dropout Layer 1 & Dropout Rate = 0.4 \\
\hline
Dense Layer 2 & Neurons = 256, Activation = tanh \\
\hline
Dropout Layer 2 & Dropout Rate = 0.4 \\
\hline
Dense Layer 3 & Neurons = 128, Activation = tanh \\
\hline
Dropout Layer 3 & Dropout Rate = 0.4 \\
\hline
Dense Layer 4 & Neurons = 64, Activation = tanh \\
\hline
Dropout Layer 4 & Dropout Rate = 0.4 \\
\hline
Output Layer & Neurons = 8, Activation = Softmax \\
\hline
\end{tabular}
\label{tab:ann_model_architecture}
\end{center}
\end{table}

\item \textbf{Applying Deep Learning Techniques:} In this context, the  ANN, RNN and CNN architecture used for deep learning are shown in Table  \ref{tab:cnn_model_architecture}, \ref{tab:rnn_model_architecture}
and \ref{tab:ann_model_architecture}. We have used the same ANN, RNN and CNN architecture that is used in \textbf{Phase 4}.

\end{enumerate}

\section{Implementation and Results}\label{sec:cdimple}
In this section, we discuss the experimental setup and analysis of the obtained results.
\subsection{\textbf{Experimental Setup}}
Our experimental setup aimed at evaluating the effectiveness of deep-learning techniques in malware classification. It consists of the following components:

\begin{enumerate}
    \item \textbf{Analysis Environment}
    \begin{itemize}
        \item \textbf{Host OS:} We used Ubuntu 18.04 LTS on a machine with an Intel i7 processor, 8GB RAM, and a 10TB HDD.
    \end{itemize}
    
    \item \textbf{Cuckoo Sandbox}
    \begin{itemize}
        \item \textbf{Version:} Cuckoo Sandbox 2.0.7 was employed for malware analysis on the Ubuntu host.
    \end{itemize}
    
    \item \textbf{Malware Samples}
    \begin{itemize}
        \item We collected two lakhs of a diverse set of malware samples representing seven categories: adware, backdoor, downloader, spyware, trojan, virus, and worm.
    \end{itemize}
    
    \item \textbf{Windows 10 Environment}
    \begin{itemize}
        \item A separate Windows 10 environment was used with an Intel i7 processor, 128GB RAM, and 5TB storage to collect and analyse dynamic analysis reports from Cuckoo Sandbox.
    \end{itemize}
  
\end{enumerate}
\subsection{\textbf{Experimental Results}}\label{sec:exper_result}
\begin{table}[htbp]
\caption{Dataset 1 (collected during January to June 2023)}
\begin{center}
\begin{tabular}{@{\extracolsep{\fill}}lrrrr@{}}
\hline
\textbf{No.} & \textbf{\textit{Types}}& \textbf{\textit{Test Sample}}& \textbf{\textit{Train Sample}}& \textbf{\textit{Total Sample}} \\
\hline\hline
1& Adware &406 & 1580 &\textbf{1986} \\
\hline
2& Backdoor &123 & 551  &\textbf{674}\\
\hline
3& Downloader &495 & 2002 &\textbf{2497}\\
\hline
4& Spyware &190 & 756 &\textbf{946}\\
\hline
5& Trojan &695 & 2873 &\textbf{3568}\\
\hline
6& Worm &277 & 1080 &\textbf{1357}\\
\hline
7& Virus &500 & 1892 &\textbf{2392}\\
\hline
8& Benign &1724 & 6910 &\textbf{8634}\\
\hline\hline
&\multicolumn{1}{r}{\textbf{Total}} & \textbf{4410} &\textbf{17644} &\textbf{22054}\\
\hline
\end{tabular}
\label{tab7:mal_data}
\end{center}
\end{table}
For the malware classification experiment, we used a total of 22054 samples, including 4410 test samples and 17644 train samples, across various malware types as well as Benign samples as shown in Table \ref{tab7:mal_data} as dataset (Dataset 1). This dataset was collected from VirusShare from January to June 2023 release and we tested our deep-learning model without concept drift.

\begin{figure}[ht]%
    \centering
    {{\includegraphics[width=8.1cm]{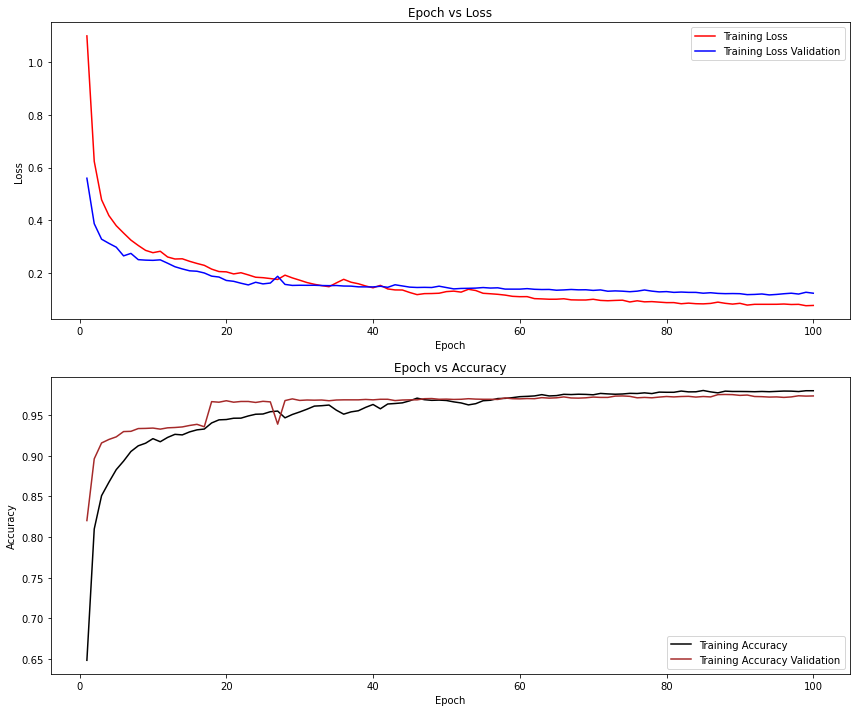} }}%
    \centering
    \caption{Epoch vs Loss and Epoch vs Accuracy for ANN}
    \label{fig:Annloss}%
\end{figure}

Table \ref{tab:ANNepochs} and Figure \ref{fig:Annloss} illustrate the performance of the Artificial Neural Network (ANN) model across different epochs during training. Initially, the model exhibits a high loss of 1.1005 and a low accuracy of 0.5595 in the first epoch. However, as the number of epochs increases, the loss steadily decreases, reaching 0.0761 by the 100$^{th}$ epoch, accompanied by a significant increase in accuracy to 0.9799. The validation metrics mirror this trend, with the validation accuracy reaching 0.9735 at the 100$^{th}$ epoch.
\begin{table}[h!]
\caption{Loss and Accuracy for Different Epochs of ANN}
\begin{center}
\begin{tabular}{@{\extracolsep{\fill}}lrrrrr@{}}
 \hline
        \textbf{No.} & \textbf{Epoch} & \textbf{Loss} & \textbf{Val Loss} & \textbf{Accuracy} & \textbf{Val Accuracy} \\
\hline \hline
1 & 1 & 1.1005 & 0.5596 & 0.6485 & 0.8203 \\
\hline
2 & 20 & 0.1959 & 0.1678 & 0.9461 & 0.9658 \\
\hline
3 & 40 & 0.1524 & 0.149 & 0.9577 & 0.9694 \\
\hline
4 & 60 & 0.1096 & 0.14 & 0.973 & 0.9703 \\
\hline
5 & 80 & 0.0869 & 0.1258 & 0.978 & 0.9723 \\
\hline
6 & 100 & 0.0761 & 0.1226 & 0.9799 & 0.9735 \\
\hline
    \end{tabular}
    \label{tab:ANNepochs}
    \end{center}
\end{table}

\begin{figure}[H]%
    \centering
    {{\includegraphics[width=8.1cm]{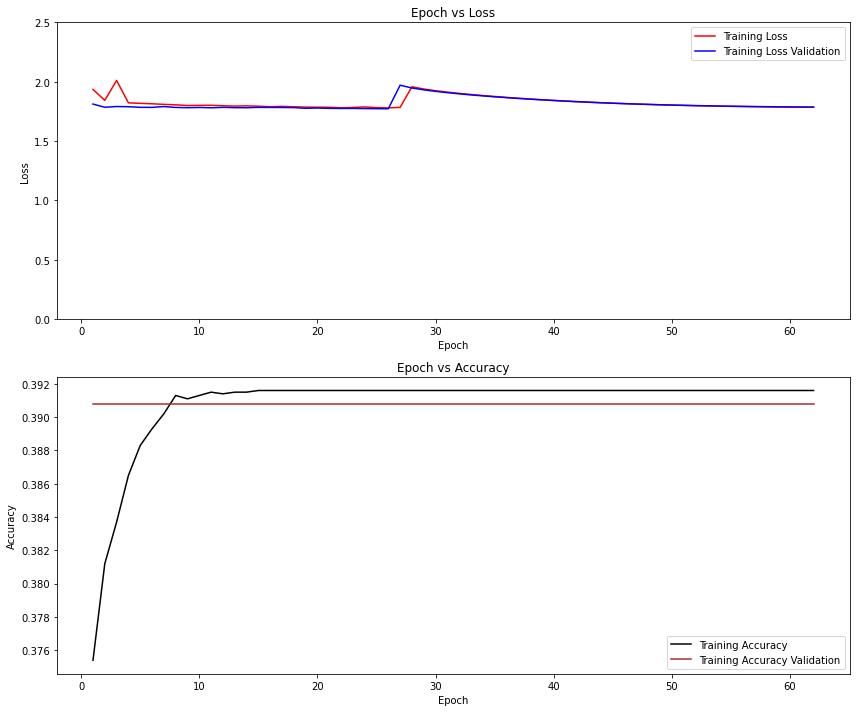} }}%
    \centering
    \caption{Epoch vs Loss and Epoch vs Accuracy for RNN}
    \label{fig:Rnnloss}%
\end{figure}

In contrast, the Recurrent Neural Network (RNN) model demonstrates different behaviour. As shown in Table \ref{tab:RNNepochs} and Figure \ref{fig:Rnnloss}, the RNN model maintains a relatively constant and high loss, ranging from 1.9359 in the first epoch to 1.6867 by the 100$^{th}$ epoch, along with a consistent accuracy of 0.3916. Similarly, the validation metrics remain stagnant, with the validation loss at 1.8181 and validation accuracy at 0.3908 throughout training.

\begin{figure}[H]%
    \centering
    {{\includegraphics[width=8.1cm]{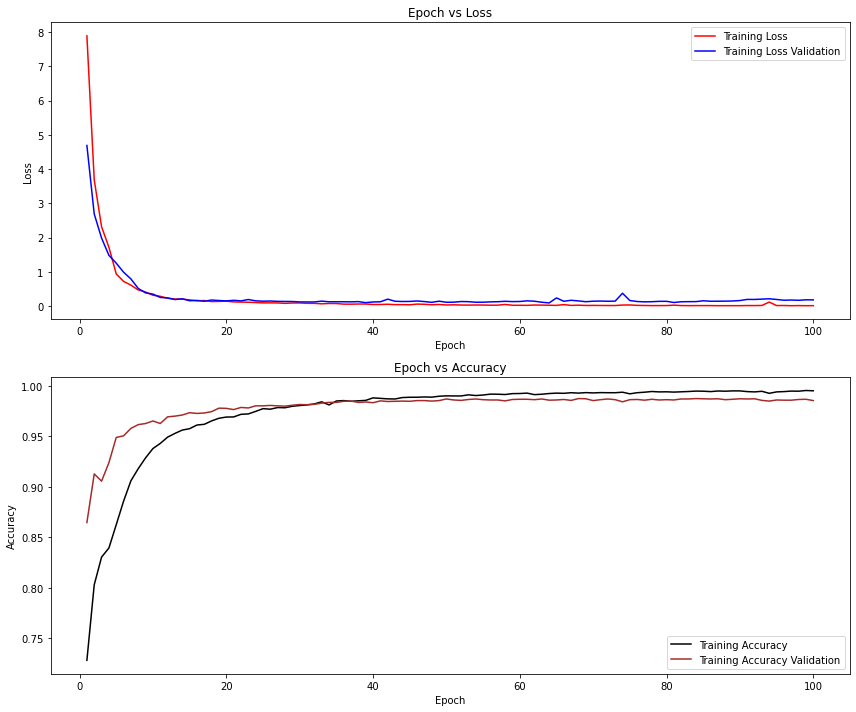} }}%
    \centering
    \caption{Epoch vs Loss and Epoch vs Accuracy for CNN}
    \label{fig:Cnnloss}%
\end{figure}

The Convolutional Neural Network (CNN) model shows the most promising results among the three models. As depicted in Table \ref{tab:CNNepochs} and Figure \ref{fig:Cnnloss}, the CNN model starts with a high loss of 7.8877 in the first epoch but experiences a rapid decrease with increasing epochs, reaching 0.0161 by the 100$^{th}$ epoch. The accuracy also improves significantly, starting at 0.7281 and reaching 0.995 by the 100$^{th}$ epoch. The validation metrics follow a similar pattern, with the validation accuracy peaking at 0.9853 by the 100$^{th}$ epoch.
\begin{table}[h!]
\caption{Loss and Accuracy for Different Epochs of RNN}
\begin{center}
\begin{tabular}{@{\extracolsep{\fill}}lrrrrr@{}}
 \hline
\textbf{No.} & \textbf{Epoch} & \textbf{Loss}   & \textbf{Val Loss} & \textbf{Accuracy} & \textbf{Val Accuracy} \\ \hline \hline
    1 &1     & 1.9359 & 1.812    & 0.3754   & 0.3908       \\ \hline
    2 &20    & 1.785  & 1.7761   & 0.3916   & 0.3908       \\ \hline
    3 &40    & 1.8376 & 1.8367   & 0.3916   & 0.3908       \\ \hline
    4 &60    & 1.7867 & 1.7881   & 0.3916   & 0.3908       \\ \hline
    3 &80    & 1.7376 & 1.7967   & 0.3916   & 0.3908       \\ \hline
    4 &100    & 1.6867 & 1.8181   & 0.3916   & 0.3908       \\ \hline
    \end{tabular}
    \label{tab:RNNepochs}
    \end{center}
\end{table}

To summarize, we can say that the CNN model outperforms the ANN and RNN models in terms of achieving lower loss and higher accuracy. The ANN model also shows promising results, with a steady improvement in performance with more epochs. However, the RNN model struggles, showing minimal improvement in loss and accuracy throughout training. These results suggest that CNN model is most suitable for this task, followed by ANN model, while RNN model may not be well-suited for this problem, i.e., classifying the malwares without concept drift.

Table \ref{tab:newdata} shows the performance metrics obtained after applying Artificial Neural Network (ANN), Convolutional Neural Network (CNN), and Recurrent Neural Network (RNN) on a new dataset of malware, downloaded from VirusShare and released between January and June 2023. It displays the maximum accuracy and loss achieved by each model.

\begin{table}[h!]
\caption{Loss and Accuracy for Different Epochs of CNN}
\begin{center}
\begin{tabular}{@{\extracolsep{\fill}}lrrrrr@{}}
 \hline

\textbf{No.} & \textbf{Epoch} & \textbf{Loss} & \textbf{Val Loss} & \textbf{Accuracy} & \textbf{Val Accuracy} \\ \hline \hline
1 & 1 & 7.8877 & 4.6905 & 0.7281 & 0.8645 \\ \hline
2 & 20 & 0.1254 & 0.1722 & 0.9691 & 0.9764 \\ \hline
3 & 40 & 0.0505 & 0.1292 & 0.9875 & 0.985 \\ \hline
4 & 60 & 0.0248 & 0.157 & 0.9927 & 0.9866 \\ \hline
5 & 80 & 0.0312 & 0.1081 & 0.9937 & 0.9859 \\ \hline
6 & 100 & 0.0161 & 0.1846 &0.995  & 0.9853 \\ \hline
    \end{tabular}
    \label{tab:CNNepochs}
    \end{center}
\end{table}

\begin{table}[h!]
\caption{Performance metrics obtained using ANN, CNN, and RNN on Dataset 1 (Without concept drift handling)}
\begin{center}
\begin{tabular}{@{\extracolsep{\fill}}lrrrrrr@{}}
 \hline

\textbf{DL} & \textbf{Epoch} & \textbf{Loss} & \textbf{Val Loss} & \textbf{Accuracy} & \textbf{Validation} \\
\textbf{Tech.} & & & & & \textbf{Accuracy}\\ \hline
\hline
ANN & 100 &0.0761  &0.1226  &0.9799  &0.9735  \\ \hline
CNN &  100 &0.0161  &0.1846  &0.9950  &0.9853  \\ \hline
RNN &  100 &1.6867  &1.8181 &0.3916  &0.3908  \\ \hline
\end{tabular}
\label{tab:newdata}
\end{center}
\end{table}

Table \ref{tab7:mal_data2} shows the dataset (Dataset 2) used for the concept drift handling approach. It contains information about different types of samples in the dataset, including Adware, Backdoor, Downloader, Spyware, Trojans, Worms, Virus, and Benign files. For each type, the table lists the number of test samples, train samples, and the total number of samples. The total number of samples in the dataset is 22,000, with 4,400 test samples and 17,600 train samples. This dataset was collected from VirusShare from the January to May 2020 release.

\begin{table}[htbp]
\caption{Dataset 2 (collected during January to May 2020)}
\begin{center}
\begin{tabular}{@{\extracolsep{\fill}}lrrrr@{}}
\hline
\textbf{No.} & \textbf{\textit{Types}} & \textbf{\textit{Test Sample}} & \textbf{\textit{Train Sample}} & \textbf{\textit{Total Sample}} \\
\hline\hline
1 & Adware & 400 & 1600 & \textbf{2000} \\
\hline
2 & Backdoor & 132 & 528 & \textbf{660} \\
\hline
3 & Downloader & 480 & 1920 & \textbf{2400} \\
\hline
4 & Spyware & 180 & 720 & \textbf{900} \\
\hline
5 & Trojan & 700 & 2800 & \textbf{3500} \\
\hline
6 & Worm & 280 & 1120 & \textbf{1400} \\
\hline
7 & Virus & 520 & 2080 & \textbf{2600} \\
\hline
8 & Benign & 1720 & 6880 & \textbf{8600} \\
\hline\hline
& \multicolumn{1}{r}{\textbf{Total}} & \textbf{4400} & \textbf{17600} & \textbf{22000} \\
\hline
\end{tabular}
\label{tab7:mal_data2}
\end{center}
\end{table}

\begin{table}[h!]
\caption{Performance metrics obtained using ANN, CNN, and RNN  on Dataset 2 (Without concept drift handling)}
\begin{center}
\begin{tabular}{@{\extracolsep{\fill}}lrrrrr@{}}
 \hline

\textbf{DL} & \textbf{Epoch} & \textbf{Loss} & \textbf{Val Loss} & \textbf{Accuracy} & \textbf{Validation} \\
\textbf{Tech.} & & & & & \textbf{Accuracy}\\ \hline
\hline
ANN & 100 &0.6250  &8.0510  &0.9112  &0.8946  \\ \hline
CNN &  100 &0.1931  &0.2864  &0.9250  &0.9430  \\ \hline
RNN &  100 &2.1200  &2.5670  &0.3429  &0.3552  \\ \hline
\end{tabular}
\label{tab:olddata}
\end{center}
\end{table}

\begin{table}[h!]
\caption{Performance metrics obtained using ANN, CNN, and RNN on Dataset 2 (With concept drift handling)}
\begin{center}
\begin{tabular}{@{\extracolsep{\fill}}lrrrrr@{}}
 \hline

\textbf{DL} & \textbf{Epoch} & \textbf{Loss} & \textbf{Val Loss} & \textbf{Accuracy} & \textbf{Validation} \\
\textbf{Tech.} & & & & & \textbf{Accuracy}\\ \hline
\hline
ANN & 100 &0.6314  &6.8510  &0.9343  &0.9184  \\ \hline
CNN &  100 &0.1429  &0.2422  &0.9314  &0.9459  \\ \hline
RNN &  100 &1.9150  &2.1563  &0.3501  &0.3558  \\ \hline
\end{tabular}
\label{tab:olddatacd}
\end{center}
\end{table}

 In contrast, Table \ref{tab:olddata} presents the performance of these models on an older dataset from VirusShare, released between January and May 2020, highlighting the  decrease in validation accuracy as well as the increase in validation loss. Moreover, Table \ref{tab:olddatacd}  demonstrates the impact of concept drift on the same old dataset (Dataset 2), showing how the models perform when trained on data that has evolved since its original release. These findings suggest that utilising concept drift in the malware dataset can lead to improved testing accuracy and reduced loss, indicating the effectiveness of the concept drift malware classifier.
\section{Comparison with Related work}\label{sec:cdcom}

 Our work focuses on improving malware classification using NLP-based $n$-gram API sequence coupled with deep learning and concept drift handling with genetic algorithms. Since we utilise a unique dataset, we lack direct comparisons with existing state-of-the-art techniques. Our approach harnesses the power of genetic algorithms, deep learning and  $n$-gram analysis, offering a distinctive perspective on malware detection that can handle concept drift. In the absence of any directly related work, we compare our work with some closely related work.

 García et al. \cite{garcia2023effectiveness} proposed a technique to assess the effectiveness of transfer learning (TL) methods for malware classification in the presence of concept drift, focusing on various time periods and learning scenarios. The study utilised five TL algorithms—TrAda, CORAL, DAE, DTS, and TIT— and evaluated their effectiveness in handling concept drift in malware classification. TrAda, CORAL, and DAE were identified as the most effective TL algorithms, consistently achieving Matthews correlation coefficients (MCC) greater than 0.9. Among the machine learning (ML) algorithms, Random Forest (RF) demonstrated competitive performance, especially in the inductive TL setting. The dataset used in the study was not explicitly disclosed by the authors, making the exact source of the malware and benign dataset unclear.

Fernando and Komninos \cite{fernando2024fesad} introduced the feSAD ransomware detection framework, which leverages machine learning to adapt to concept drift. The framework aims at enhancing ransomware detection rates by calibrating drift thresholds and identifying abnormal drift samples. Compatible with various machine learning algorithms, it has demonstrated the effectiveness in detecting ransomware amidst concept drift and can be tailored for different malware types. Experimental results, including 720 ransomware samples and 2000 benign samples from the Elderan dataset, show high detection rates and precision, especially with the Random Forest algorithm. Fernando and Komninos \cite{fernando2024fesad} highlighted the Random Forest's stability in ransomware detection. However, they noted that in test batch 2, a high volume of ransomware samples caused abnormal concept drift, leading to reduced detection rates and statistical drift, suggesting system retraining to avoid detection decline.

\begin{table*}[ht]

\caption{Comparison with some related work}
\begin{center}
\begin{tabular}{@{\extracolsep{\fill}}lrcrrrrr@{}}
 \hline
 \textbf{S No.} &\textbf{Authors} &\textbf{$n$-gram API} &\textbf{Concept Drift} &\textbf{Dataset Source} &\textbf{Dataset}  &\textbf{Features }\\
 & &\textbf{Seq. used} &\textbf{Tech. used} & &\textbf{Size} &\textbf{Used}\\
 \hline\hline

1 &García et al. \cite{garcia2023effectiveness}                 & \ding{55}                          &\textbf{Transfer learning} & Not Disclosed &9196  &APIs, signatures, \\    & & & & & &and network \\
\hline
2 &Fernando and                 & \ding{51}                          &\textbf{Genetic Algorithms} & Elderan dataset & 2720  & API Calls\\
&Komninos\cite{fernando2024fesad} &&&&&\\
\hline

3 &Proposed Work                         & \ding{51}            &\textbf{Genetic Algorithms} & VirusShare  &22054  & API Calls \\
\hline
\end{tabular}
\label{tab:com}
\end{center}
\end{table*}

\section{Threats to Validity}\label{sec:threat}
There are some potential threats to the validity of the proposed model and its results. We discussed them below.

\textbf{Internal Validity}
\begin{itemize}
    \item \textit{Algorithm Performance}: The effectiveness and the performance of the genetic algorithm approach for concept drift handling could be influenced by the specific parameters and configurations chosen, which may impact the results.

\end{itemize}

\textbf{External Validity}
\begin{itemize}

    \item \textit{Concept Drift Representation}: The concept drift handling approach may not fully capture the complexity and dynamics of concept drift in real-world malware datasets, affecting the effectiveness of the proposed approach in practical scenarios.
\end{itemize}

\textbf{Construct Validity}
\begin{itemize}
    
    \item \textit{Model Architecture}: The specific architecture of the neural network used for classification may not be optimal for handling concept drift or may not fully leverage the additional features added through genetic algorithms.
\end{itemize}

\textbf{Conclusion Validity}
\begin{itemize}
    \item \textit{Evaluation Metrics}: The evaluation metrics used (e.g., accuracy, loss) may not fully capture the effectiveness of the approach in handling concept drift and distinguishing between malware and benign samples, potentially leading to biased conclusions.
\end{itemize}

Addressing these threats involves conducting thorough experiments with diverse datasets, carefully selecting parameters and configurations for the genetic algorithm and neural network, and considering the implications of concept drift on the model's performance and generalizability.

\section{Conclusion and Future Work}\label{sec:conclusion}
In this paper, we presented a deep learning approach for dynamic PE malware analysis, focusing on handling concept drift. Our model includes the phases of preprocessing, NLP processing, feature selection, malware classification, and concept drift handling.

We used API call frequency features for initial selection and added 10,500 features using genetic algorithms, improving the model's ability to distinguish malware from benign samples and handle concept drift.

Our neural network has an input layer with 88,972 neurons, three hidden layers, and an output layer with eight neurons for multi-class classification. ReLU activation and Adam optimizer were used for training, resulting in high accuracy and low loss.

We calculated fitness scores for 101,248 new mutants, selecting the top 10,500 for feature augmentation, increasing the feature corpus by approximately 1\%.

Experimental results showed significant improvements in classification accuracy and loss reduction across a number of epochs and datasets, indicating the potential of our approach for enhancing malware detection and classification in dynamic PE environments, with concept drift handling.

\section*{Scope for Future Research}
Below we present some future scope of our research work.
\begin{itemize}
    \item Explore the application of other deep learning approaches, such as Long Short Term Memory Networks (LSTMs) and Generative Adversarial Networks (GANs), to broaden the scope of the research and potentially uncover new insights.
    \item Investigate the use of genetic algorithm approaches for feature selection and feature creation to enhance the robustness and accuracy of the models.
    \item Experiment with transfer learning techniques to leverage pre-trained models for malware classification tasks, potentially improving performance and reducing training time.
    \item Conduct cross-domain analysis by applying the developed models to different types of malware datasets, assessing their generalizability and adaptability to diverse malware threats.
\end{itemize}

\section{Code and Dataset Availability}
The code for this work is available on GitHub at the following link: \url{https://github.com/bishwajitprasadgond/MalClassCD}. For access to the dataset used in this research, please send a request via email to \url{bishwajitprasadgond@gmail.com}.
 \section{Declaration of competing interest}
The authors declare that they have no known competing financial
interests or personal relationships that could have appeared to influence
the work reported in this paper.
\bibliographystyle{unsrt}
\bibliography{reference}
\end{document}